\pdfoutput=1

\documentclass[11pt]{article}

\usepackage[final]{acl}
\usepackage{natbib}

\usepackage{times}
\usepackage{latexsym}

\usepackage[T1]{fontenc}

\usepackage[utf8]{inputenc}

\usepackage{soul} 

\usepackage{microtype}

\usepackage{dirtytalk}
\usepackage{booktabs}
\usepackage{graphicx}
\usepackage{subcaption}

\usepackage{amsmath}
\title{Large Reasoning Models are not thinking straight: on the unreliability of thinking trajectories}

\author{
  \textbf{Jhouben Cuesta-Ramirez\textsuperscript{1}\textsuperscript{*}}, 
  \textbf{Samuel Beaussant\textsuperscript{1}\textsuperscript{*}}, 
  \textbf{Mehdi Mounsif\textsuperscript{1}\textsuperscript{*}} \\
  \textsuperscript{1}Akkodis Research \\
  \textsuperscript{*}Equal contributions \\
  \texttt{\{Jhouben-Janyk.CUESTA-RAMIREZ, samuel.beaussant, mehdi.mounsif\}@akkodis.com}
}
\begin{document}
\maketitle
\begin{abstract}

Large Language Models (LLMs) trained via Reinforcement Learning (RL) have recently achieved impressive results on reasoning benchmarks. Yet, growing evidence shows that these models often generate longer but ineffective chains of thought (CoTs), calling into question whether benchmark gains reflect real reasoning improvements. We present new evidence of \textbf{overthinking}, where models disregard correct solutions even when explicitly provided, instead continuing to generate unnecessary reasoning steps that often lead to incorrect conclusions. Experiments on three state-of-the-art models using the AIME2024 math benchmark reveal critical limitations in these models’ ability to integrate corrective information, posing new challenges for achieving robust and interpretable reasoning.
\end{abstract}

\section{Introduction}
Recent advancements have seen LLMs achieve impressive scores on diverse linguistic and cognitive benchmarks, suggesting significant progress in their reasoning capabilities \cite{havellmsadvancedenoughachallengingproblemsolvingbenchmarkforlargelanguagemodels}. However, this performance often degrades substantially when encountering out-of-distribution (OOD) tasks  revealing limitations in their reliability and adaptive reasoning \cite{selfconsistencyimproveschainofthoughtreasoninginlanguagemodels}.

Strategies such as enhanced test-time inference budgets offer potential mitigation for OOD challenges \cite{advancinglanguagemodelreasoningthroughreinforcementlearningandinferencescaling} but a more focused approach involves specialized reasoning models trained via reinforcement learning (RL) with verifiable rewards \cite{critiqueoutloudrewardmodels}. This enables backtracking and self-revision, significantly outperforming standard LLMs. The CoT embodying these strategies are themselves a subject of intense study, offering potential blueprints for enhancing reasoning in smaller models \cite{treeofthoughtsdeliberateproblemsolvingwithlargelanguagemodels}.


Nevertheless, concerns remain regarding the authenticity of these CoTs. Prior work \cite{planandsolvepromptingimprovingzeroshotchainofthoughtreasoningbylargelanguagemodels} has shown that cold-start RL can produce unconventional CoTs (often with unusual syntax or mixed languages) that still lead to correct answers. This raises doubts about whether these CoTs represent genuine reasoning or merely environment-specific artifacts anthropomorphized by researchers \cite{interpretablecontrastivemontecarlotreesearchreasoning}. Such phenomena may result from RL incentives that encourage retrieving known solutions from the base model, rather than fostering new cognitive skills \cite{yue2025does}. Recent evidence even suggests that disabling explicit reasoning processes in LRMs can still achieve state-of-the-art results if sampling budgets are sufficiently increased \cite{ma2025reasoning}. 

We present new evidence suggesting that CoTs may not meaningfully support models' reasoning processes. Extending prior work \cite{underthinking}, we explore the \textbf{underthinking} phenomenon, where models prematurely abandon promising reasoning paths maybe due to inability to correctly identify a valid solution. We find that even explicitly injecting ground-truth solutions into reasoning sequences often fails, with models instead pursuing incorrect reasoning. This issue closely relates to \textbf{overthinking} \cite{overthinking}, where models generate excessive, unproductive reasoning tokens. In this preliminary work, we demonstrate empirically that even state-of-the-art reasoning models are not apart from this issue with two main contributions: Demonstrating that LLMs often disregard externally provided correct solutions, producing redundant reasoning steps and Highlighting fundamental limitations in the integration of corrective information by models, revealing critical flaws in current reasoning dynamics.

\section{Related Work}
\label{sec:relwork}

Following the seminal work of \cite{wei2022chain}, suggesting that LLM's reasoning capabilities can be enhanced \textit{via} Chain-of-Thought (CoT), numerous works have explored prompt augmentations \cite{treeofthoughtsdeliberateproblemsolvingwithlargelanguagemodels, selfconsistencyimproveschainofthoughtreasoninginlanguagemodels}, specific scaffolding \cite{bairi2023codeplanrepositorylevelcodingusing, inductive-thinking, LLMmctsAutomaticHeuristicDesign}. Concurrently, research has also highlighted that expanding the sampling budget can lead to further performance gains \cite{wang2024mixtureofagentsenhanceslargelanguage, tian2024selfimprovementllmsimaginationsearching}. For instance, \cite{LargeLanguageMonkeys} observed a strong correlation between sampling budget and benchmark performance, indicating that models may have internalized much more knowledge than what is revealed through single-shot forward pass. However, while these methods can enhance performance, their reliability often rests on the capabilities of the verifier. This challenge aligns naturally with the reinforcement learning (RL) framework, where algorithms such as \cite{ppo, grpo} leverage ground truth rewards as training signals to optimize LLMs \cite{o1_syscard, o3o4_syscard, deepseekR1, Tulu}.

Despite initial competitive metrics and hill-climbing trends \cite{deepscaler2025}, fundamental questions about LLMs' planning capabilities persist \cite{llm-modulo, lrm-modulo}, prompting skepticism about their CoTs' relevance \cite{underthinking, overthinking}. RL optimization often leads to overfitting, unreadable CoTs, or anthropomorphic strategies \cite{qwq-32b-preview}, emphasizing the need for robust reasoning and interpretability \cite{interpretabletranscoder, sparseencoder} to handle shifted distributions in inference \cite{jailbreakingalignedllms}. Recent works \cite{llm-circuit, biology-llm} have identified interpretable internal circuits within transformer models, but scaling these insights remains challenging \cite{gemmaSAE}. Evidence of sycophancy and divergence between textual outputs and internal activations further complicates system trustworthiness. Our work fundamentally challenges the assumed relationship between CoT length and performance by demonstrating multiple models' inability to recognize correct answers even when explicitly presented. Along with recent studies \cite{S1RL, demystifyinglongchainofthoughtreasoning, nothinking, optimizingforpassatk} these results reveal a more complex dynamic influenced by training dynamics and task complexity, underscoring the need for a deeper understanding of LLM reasoning processes.

\flushbottom
\begin{figure*}[!t]
    \centering
    \begin{subfigure}{0.49\textwidth}
        \centering
        \includegraphics[width=\linewidth]{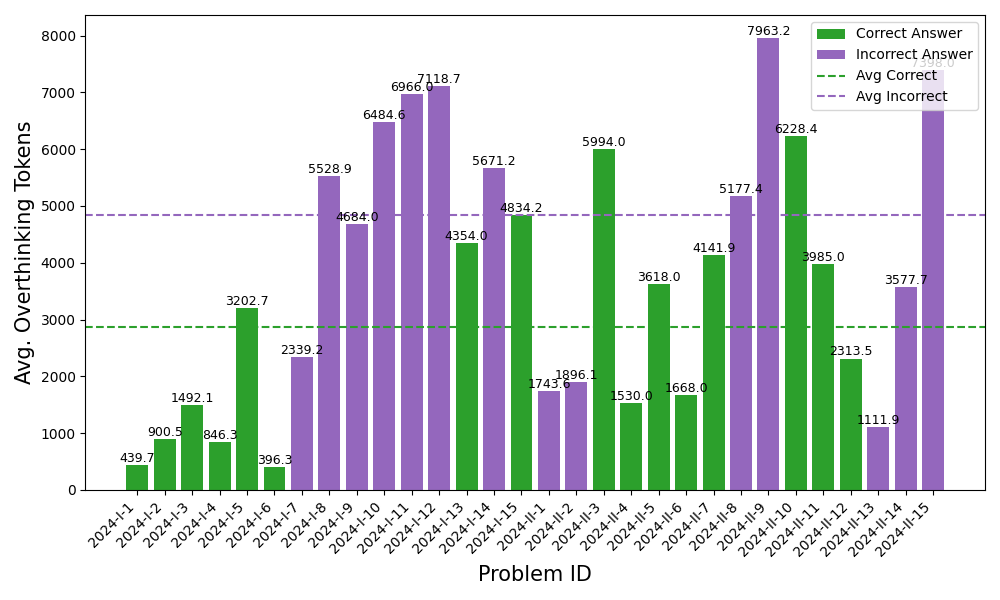}
        \caption{Llama-70b}
        \label{fig:subfig1}
    \end{subfigure}
    \hfill
    \begin{subfigure}{0.49\textwidth}
        \centering
        \includegraphics[width=\linewidth]{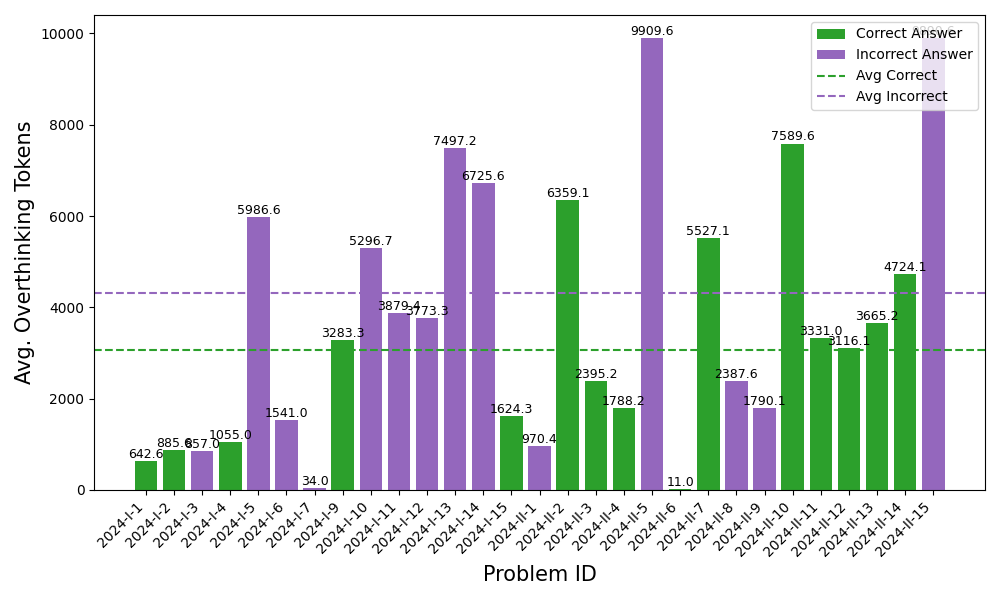}
        \caption{Qwen-7b}
        \label{fig:subfig2}
    \end{subfigure}
    \hfill
    \begin{subfigure}{0.49\textwidth}
        \centering
        \includegraphics[width=\linewidth]{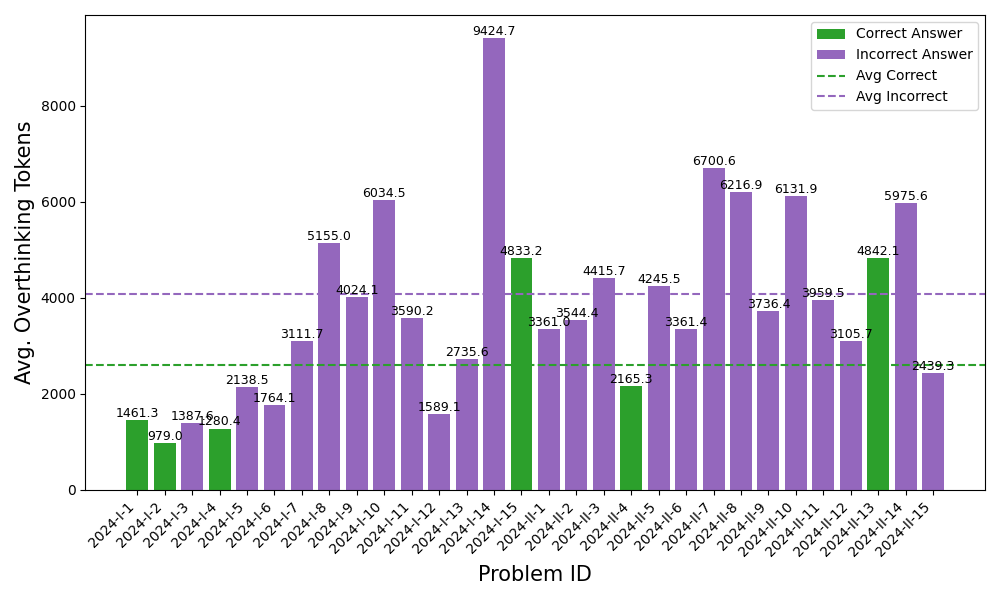}
        \caption{DeepScalR-1.5b}
        \label{fig:subfig3}
    \end{subfigure}
    \hfill
    \caption{Averaged number of \textit{overthinking} tokens generated for each problem from AIME2024. Amount of \textit{overthinking} tokens is problem-dependent but all four models overthink. The correct/incorrect coloring is based on whether the problem was successfully solved on the first try.
    }
    \label{fig:main}
\end{figure*}

\section{Results and Discussion}
\label{sec:results}
We conduct our experiments on three different models. LLaMA 70B and Qwen7B were distilled from R1 \cite{deepseekR1}, while DeepScalR1.5B \cite{deepscaler2025} was trained via RL. Our reasoning analysis focus on the AIME2024 dataset, a challenging set of mathematical problems.

\subsection{Baseline Thoughts Generation}
Baseline solutions are generated by providing the models with a standard system prompt and problem statement. The LLM then produces CoTs enclosed in $\text{<think>}$ tags before delivering a final answer, spontaneously exploring multiple hypotheses and engaging in self-reflection. Individual reasoning thoughts are identified using a simple heuristic based on specific keywords, with $\textit{Alternatively}$ and $\textit{Wait}$ serving as thought separators. We denote the reasoning trace up to the $t^\text{th}$ thought as $\mathcal{R}{[:t]}$, which includes the system prompt and problem statement. For $t=0$, $\mathcal{R}{[:0]}$ contains only the prompt and statement; for $t=T$, $\mathcal{R}_{[:T]}$ represents the full reasoning trajectory. This notation enables the analysis of reasoning at any intermediate step $t$, supporting counterfactual studies.

\subsection{Ground Truth Integration}

\begin{figure*}[!t]
    \centering
    \begin{subfigure}{0.45\textwidth}
        \centering
        \includegraphics[width=1.05\linewidth]{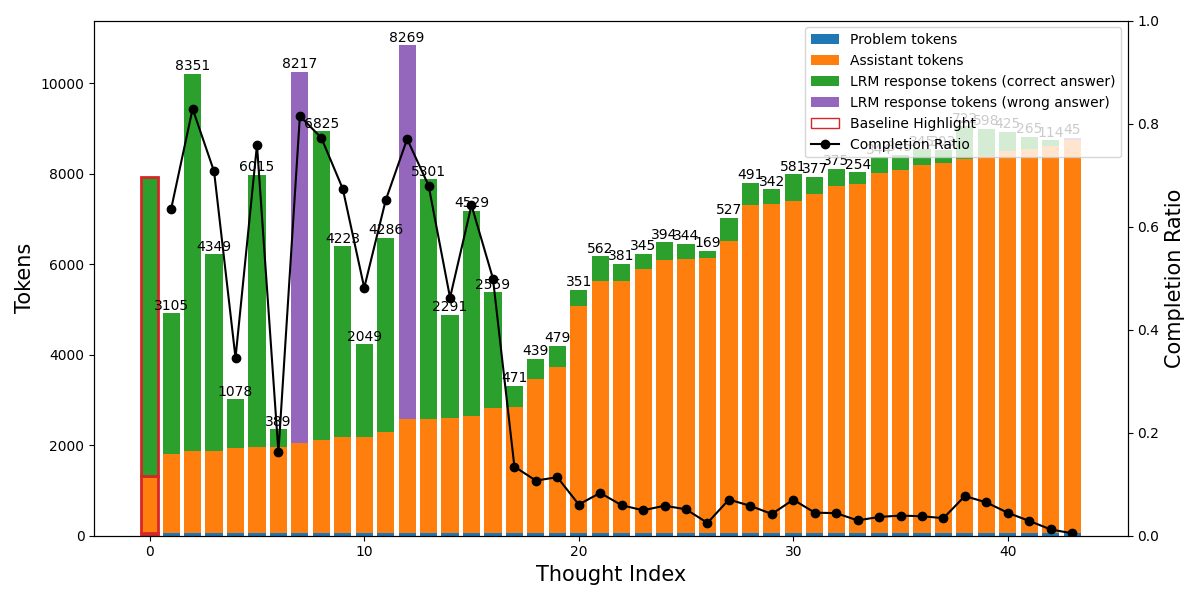}
        \caption{2024-I-5}
        \label{fig:subfig_I-5}
    \end{subfigure}
    \hfill
    \begin{subfigure}{0.45\textwidth}
        \centering
        \includegraphics[width=1.05\linewidth]{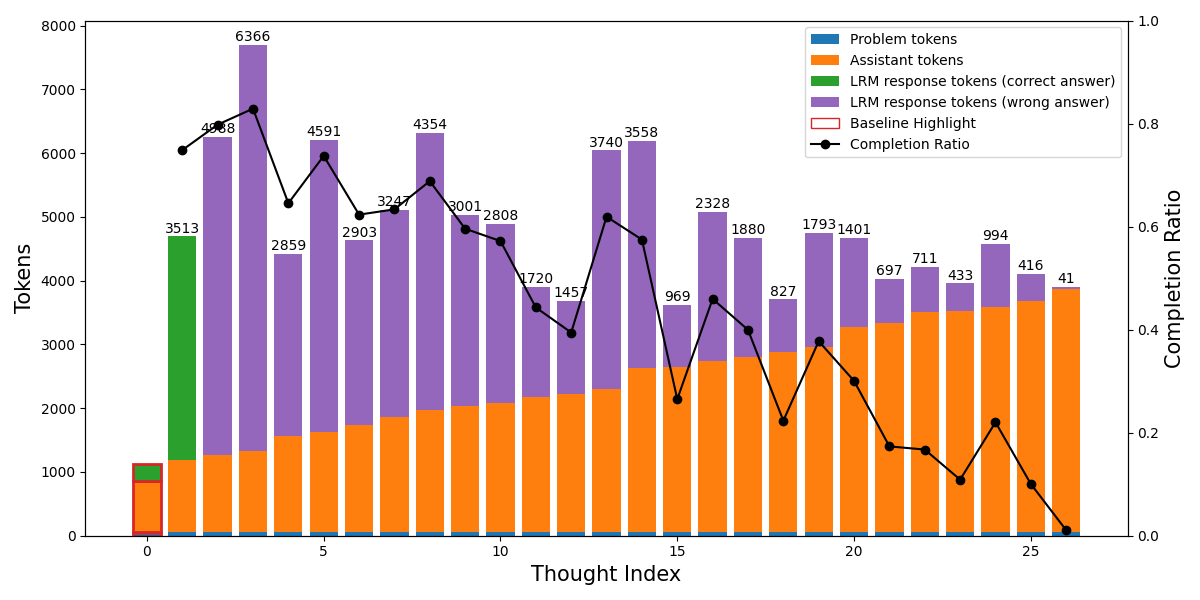}
        \caption{2024-II-5}
        \label{fig:subfig_II-15}
    \end{subfigure}
    \caption{Illustration of the \textit{overthinking} issue on two problems from AIME2024 with DeepScalR1.5B. $P_t$ is pictured in orange (assistant tokens) while "LRM response" represents the completion tokens with the coloring indicating wether the model predicted the correct answer or not. The completion ratio indicates the amount of completion tokens generated in proportion to the total amount of tokens.}
    \label{fig:overthinking_single_sub_thought}
\end{figure*}

We investigate the impact of integrating the ground truth solution (reasoning $+$ answer) after each baseline thoughts, assessing whether models recognize and integrate correct answers. For each problem, we concatenate the solution $\mathcal{Y}$ to the already generated thinking trace $\mathcal{R}_{[:t]}$ as if it was generated by the model itself during its reasoning process. This achieved by prefixing the solution with the relevant set of tokens $\mathcal{V}$ to mimic the model's writing style. More formally, the prompt $P_t$ sent to the model is: $P_t = \mathcal{R}_{[:t]} \oplus \mathcal{V} \oplus \mathcal{Y} ~\refstepcounter{equation}(\theequation)\label{eq:p_t}$, where $\oplus$ denotes concatenation and with $\mathcal{V} = \text{"<think> Okay, so "} \, \text{if } t = 0 $ and $\text{"Alternatively, "}$ otherwise.
Given $P_t$ for each $t \in [0,T]$, the model continues the reasoning trajectory without additional prompts or guidance. We expect the model to pick-up the ground truth solution and use it as its final answer. However, models often generate excessive additional tokens, frequently doubting or even discarding the correct solution, as illustrated in Figure \ref{fig:subfig_II-15}. Surprisingly, in this pathological case the model almost always commits to a wrong answer despite the given solution. However, this illustrative example is not an isolated case as shown In Figure \ref{fig:main}. This phenomenon emerges in all models regardless of their size and post-training method. The amount of overthinking tokens however is largely problem-dependent, with some specific problems creating more overthinking tokens than others (e.g. problem I-14 on Figure \ref{fig:main}).
\flushbottom


A rare positive case, observed only twice, is shown in Figure \ref{fig:subfig_I-5}. Here, after $t=17$, the model consistently accepted the injected solution, producing a reduced and stable number of tokens. The reasoning reflected diverse yet coherent strategies, such as applying cyclic quadrilaterals, similar triangles, and the British flag theorem which suggests that the model could have required a number of thoughts \cite{wei2022chain} before being able to understand and accept the injected truth. This is represented in the figure as a big reduction in the number of tokens generated and the completion ratio suggesting a \textit{converging number} of generated tokens, after a spiky, iterative and often \textit{divergent behavior} employing an excessive amount of tokens.

\begin{figure}
    \centering
    \includegraphics[width=\linewidth]{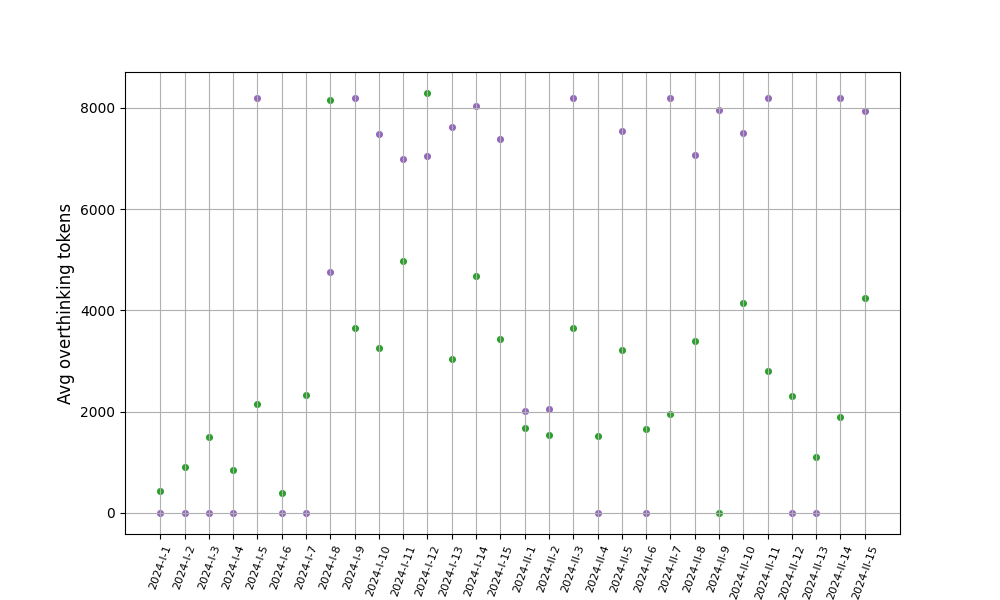}
    \caption{Averaged number of \textit{overthinking} tokens generated for each problem from AIME2024 with model LLaMA 70B. The average is conditioned to the model providing a correct answer(green) or not(purple) at the end of its reasoning.} 
    \label{fig:ciavg}
\end{figure}

We found that this divergent behavior is common in most of the problems, with an example represented in Figure \ref{fig:ciavg} where for the first four of them, the number of overthinking tokens in order to provide a bad answer is close to $0$ and to provide / accept the correct answer is lower than $2000$. Then for the rest things went out of control as this grew considerably even while accepting the truth. Which to our point of view could correspond to a clear prioritization on self-generated tokens, lack of capabilities of comprehension or simply randomly divergent CoTs.

These results reveal that reasoning models often struggle to identify valid solutions, relying on complex but fragile heuristics that do not reflect genuine reasoning. For instance, in a particularly challenging problem (II-15), none of the models confidently arrived at the correct solution, even when provided with the ground truth (Figure \ref{fig:overthinking_single_sub_thought}). When models did accept the correct answer, it was typically out of resignation or time pressure rather than true understanding, as shown by statements like, "But perhaps given the time I have, I think the initial calculation of 315 might be correct," which appeared nearly 3000 tokens after the ground truth. This shows that in some cases, the correct answer may not even exist within the model’s solution space, making it unattainable through further sampling or prompting.

This behavior likely stems from overfitting to specific patterns and spurious correlations that do not generalize. As shown in Figure \ref{fig:overthinking_single_sub_thought}, the number of overthinking tokens often decreases in proportion to the total tokens generated, though this relationship sometimes appears thresholded. Models tend to delay committing to an answer until reaching a predefined token count or detecting unknown patterns. These tendencies likely arise from RL training dynamics and poor credit assignment, where correct answers reached through faulty CoTs are mistakenly rewarded and associated with the CoT’s length or other irrelevant features. This mirrors reward hacking seen in smaller RL tasks, where optimization exploits shortcuts that maximize rewards but fail to generalize.


\section{Conclusion and Future work}

While recent advances in reasoning models have generated significant interest, their evaluation often relies on anthropomorphizing behavior, treating CoTs as reflections of human reasoning. Our findings reveal a critical limitation: current LRMs lack robust mechanisms to integrate external corrective signals. This challenges assumptions about their coherence and questions the reliability of CoTs as transparent indicators of cognition. Proxy metrics like completion length also fail to consistently reflect reasoning quality, as models frequently display both underthinking and overthinking. Our analysis shows that models often ignore provided ground-truth solutions, raising key questions: What truly drives the reasoning process, and can it be effectively guided? Models that reject external solutions are also prone to disregarding guidance, feedback, or human intervention.
 
Building on these insights our future work will focus on developing heuristics to improve model receptiveness to external suggestions or corrections. By leveraging recent advances in understanding interpretable circuits within transformer models, we aim to uncover why models tend to doubt or even reject correct answers during reasoning. This deeper understanding will inform the development of more reliable and efficient reasoning models.

\section*{Limitations}
Our study focuses solely on mathematical reasoning tasks (AIME2024), so the findings may not generalize to other domains. The set of models tested is also limited and may not capture the full diversity of training paradigms. Additionally, our method of injecting ground-truth solutions relies on heuristics that might not fully align with the models’ internal reasoning dynamics. Finally, while we provide behavioral analysis, deeper interpretability work remains for future research.

\bibliography{acl_latex}

\begin{thebibliography}{38}
\providecommand{\natexlab}[1]{#1}

\bibitem[{Ameisen et~al.(2025)Ameisen, Lindsey, Pearce, Gurnee, Turner, Chen,
  Citro, Abrahams, Carter, Hosmer, Marcus, Sklar, Templeton, Bricken,
  McDougall, Cunningham, Henighan, Jermyn, Jones, Persic, Qi, Ben~Thompson,
  Zimmerman, Rivoire, Conerly, Olah, and Batson}]{llm-circuit}
Emmanuel Ameisen, Jack Lindsey, Adam Pearce, Wes Gurnee, Nicholas~L. Turner,
  Brian Chen, Craig Citro, David Abrahams, Shan Carter, Basil Hosmer, Jonathan
  Marcus, Michael Sklar, Adly Templeton, Trenton Bricken, Callum McDougall,
  Hoagy Cunningham, Thomas Henighan, Adam Jermyn, Andy Jones, and 8 others.
  2025.
\newblock \href
  {https://transformer-circuits.pub/2025/attribution-graphs/methods.html}
  {Circuit tracing: Revealing computational graphs in language models}.
\newblock \emph{Transformer Circuits Thread}.

\bibitem[{Andriushchenko et~al.(2025)Andriushchenko, Croce, and
  Flammarion}]{jailbreakingalignedllms}
Maksym Andriushchenko, Francesco Croce, and Nicolas Flammarion. 2025.
\newblock \href {https://arxiv.org/abs/2404.02151} {{Jailbreaking Leading
  Safety-Aligned LLMs with Simple Adaptive Attacks}}.
\newblock \emph{Preprint}, arXiv:2404.02151.

\bibitem[{Ankner et~al.(2023)Ankner, Cui, Chang, and
  Ammanabrolu}]{critiqueoutloudrewardmodels}
Zachary Ankner, Mansheej Paul~Brandon Cui, Jonathan~D. Chang, and Prithviraj
  Ammanabrolu. 2023.
\newblock \href {https://arxiv.org/pdf/2408.11791} {Critique-out-loud reward
  models}.
\newblock \emph{arXiv preprint arXiv:2408.11791}.

\bibitem[{Arora and
  Singh(2023)}]{havellmsadvancedenoughachallengingproblemsolvingbenchmarkforlargelanguagemodels}
Daman Arora and Himanshu~Gaurav Singh. 2023.
\newblock \href {https://arxiv.org/pdf/2305.15074} {Have llms advanced enough?
  a challenging problem solving benchmark for large language models}.
\newblock \emph{arXiv preprint arXiv:2305.15074}.

\bibitem[{Bairi et~al.(2023)Bairi, Sonwane, Kanade, C, Iyer, Parthasarathy,
  Rajamani, Ashok, and Shet}]{bairi2023codeplanrepositorylevelcodingusing}
Ramakrishna Bairi, Atharv Sonwane, Aditya Kanade, Vageesh~D C, Arun Iyer,
  Suresh Parthasarathy, Sriram Rajamani, B.~Ashok, and Shashank Shet. 2023.
\newblock \href {https://arxiv.org/abs/2309.12499} {{CodePlan: Repository-level
  Coding using LLMs and Planning}}.
\newblock \emph{Preprint}, arXiv:2309.12499.

\bibitem[{Brown et~al.(2025)Brown, Juravsky, Ehrlich, Clark, Le, Re, and
  Mirhoseini}]{LargeLanguageMonkeys}
Bradley Brown, Jordan Juravsky, Ryan~Saul Ehrlich, Ronald Clark, Quoc~V Le,
  Christopher Re, and Azalia Mirhoseini. 2025.
\newblock \href {https://openreview.net/forum?id=0xUEBQV54B} {{Large Language
  Monkeys: Scaling Inference Compute with Repeated Sampling}}.

\bibitem[{Chen et~al.(2025)Chen, Xu, Liang, He, Pang, Yu, Song, Liu, Zhou,
  Zhang, Wang, Tu, Mi, and Yu}]{overthinking}
Xingyu Chen, Jiahao Xu, Tian Liang, Zhiwei He, Jianhui Pang, Dian Yu, Linfeng
  Song, Qiuzhi Liu, Mengfei Zhou, Zhuosheng Zhang, Rui Wang, Zhaopeng Tu,
  Haitao Mi, and Dong Yu. 2025.
\newblock \href {https://arxiv.org/abs/2412.21187} {Do not think that much for
  2+3=? on the overthinking of o1-like llms}.
\newblock \emph{Preprint}, arXiv:2412.21187.

\bibitem[{Cheng et~al.(2024)Cheng, Yang, Jiang, Wang, Huang, Li, Li, Li, Gao,
  Li, Yin, and Sun}]{inductive-thinking}
Kewei Cheng, Jingfeng Yang, Haoming Jiang, Zhengyang Wang, Binxuan Huang,
  Ruirui Li, Shiyang Li, Zheng Li, Yifan Gao, Xian Li, Bing Yin, and Yizhou
  Sun. 2024.
\newblock \href {https://arxiv.org/abs/2408.00114} {{Inductive or Deductive?
  Rethinking the Fundamental Reasoning Abilities of LLMs}}.
\newblock \emph{Preprint}, arXiv:2408.00114.

\bibitem[{Cunningham et~al.(2023)Cunningham, Ewart, Riggs, Huben, and
  Sharkey}]{sparseencoder}
Hoagy Cunningham, Aidan Ewart, Logan Riggs, Robert Huben, and Lee Sharkey.
  2023.
\newblock \href {https://arxiv.org/abs/2309.08600} {Sparse autoencoders find
  highly interpretable features in language models}.
\newblock \emph{Preprint}, arXiv:2309.08600.

\bibitem[{DeepSeek-AI(2025)}]{deepseekR1}
DeepSeek-AI. 2025.
\newblock \href {https://arxiv.org/abs/2501.12948} {Deepseek-r1: Incentivizing
  reasoning capability in llms via reinforcement learning}.
\newblock \emph{Preprint}, arXiv:2501.12948.

\bibitem[{Dunefsky et~al.(2024)Dunefsky, Chlenski, and
  Nanda}]{interpretabletranscoder}
Jacob Dunefsky, Philippe Chlenski, and Neel Nanda. 2024.
\newblock \href {https://arxiv.org/abs/2406.11944} {Transcoders find
  interpretable llm feature circuits}.
\newblock \emph{Preprint}, arXiv:2406.11944.

\bibitem[{Gao et~al.(2023)Gao, Niu, He, Xu, Liu, Liu, Hu, and
  Wen}]{interpretablecontrastivemontecarlotreesearchreasoning}
Zitian Gao, Boye Niu, Xuzheng He, Haotian Xu, Hongzhang Liu, Aiwei Liu, Xuming
  Hu, and Lijie Wen. 2023.
\newblock \href {https://arxiv.org/pdf/2410.01707} {Interpretable contrastive
  monte carlo tree search reasoning}.
\newblock \emph{arXiv preprint arXiv:2410.01707}.

\bibitem[{Hou et~al.(2025)Hou, Lv, Lu, Zhang, Li, Yao, Li, Tang, and
  Dong}]{advancinglanguagemodelreasoningthroughreinforcementlearningandinferencescaling}
Zhenyu Hou, Xin Lv, Rui Lu, Jiajie Zhang, Yujiang Li, Zijun Yao, Juanzi Li, Jie
  Tang, and Yuxiao Dong. 2025.
\newblock \href {https://arxiv.org/pdf/2501.11651} {Advancing language model
  reasoning through reinforcement learning and inference scaling}.
\newblock \emph{arXiv preprint arXiv:2501.11651}.

\bibitem[{Kambhampati et~al.(2024)Kambhampati, Valmeekam, Guan, Verma, Stechly,
  Bhambri, Saldyt, and Murthy}]{llm-modulo}
Subbarao Kambhampati, Karthik Valmeekam, Lin Guan, Mudit Verma, Kaya Stechly,
  Siddhant Bhambri, Lucas Saldyt, and Anil Murthy. 2024.
\newblock \href {https://arxiv.org/abs/2402.01817} {Llms can't plan, but can
  help planning in llm-modulo frameworks}.
\newblock \emph{Preprint}, arXiv:2402.01817.

\bibitem[{Lambert et~al.(2025)Lambert, Morrison, Pyatkin, Huang, Ivison,
  Brahman, Miranda, Liu, Dziri, Lyu, Gu, Malik, Graf, Hwang, Yang, Bras,
  Tafjord, Wilhelm, Soldaini, Smith, Wang, Dasigi, and Hajishirzi}]{Tulu}
Nathan Lambert, Jacob Morrison, Valentina Pyatkin, Shengyi Huang, Hamish
  Ivison, Faeze Brahman, Lester James~V. Miranda, Alisa Liu, Nouha Dziri, Shane
  Lyu, Yuling Gu, Saumya Malik, Victoria Graf, Jena~D. Hwang, Jiangjiang Yang,
  Ronan~Le Bras, Oyvind Tafjord, Chris Wilhelm, Luca Soldaini, and 4 others.
  2025.
\newblock \href {https://arxiv.org/abs/2411.15124} {Tulu 3: Pushing frontiers
  in open language model post-training}.
\newblock \emph{Preprint}, arXiv:2411.15124.

\bibitem[{Lieberum et~al.(2024)Lieberum, Rajamanoharan, Conmy, Smith, Sonnerat,
  Varma, Kramár, Dragan, Shah, and Nanda}]{gemmaSAE}
Tom Lieberum, Senthooran Rajamanoharan, Arthur Conmy, Lewis Smith, Nicolas
  Sonnerat, Vikrant Varma, János Kramár, Anca Dragan, Rohin Shah, and Neel
  Nanda. 2024.
\newblock \href {https://arxiv.org/abs/2408.05147} {Gemma scope: Open sparse
  autoencoders everywhere all at once on gemma 2}.
\newblock \emph{Preprint}, arXiv:2408.05147.

\bibitem[{Lindsey et~al.(2025)Lindsey, Gurnee, Ameisen, Chen, Pearce, Turner,
  Citro, Abrahams, Carter, Hosmer, Marcus, Sklar, Templeton, Bricken,
  McDougall, Cunningham, Henighan, Jermyn, Jones, Persic, Qi, Thompson,
  Zimmerman, Rivoire, Conerly, Olah, and Batson}]{biology-llm}
Jack Lindsey, Wes Gurnee, Emmanuel Ameisen, Brian Chen, Adam Pearce,
  Nicholas~L. Turner, Craig Citro, David Abrahams, Shan Carter, Basil Hosmer,
  Jonathan Marcus, Michael Sklar, Adly Templeton, Trenton Bricken, Callum
  McDougall, Hoagy Cunningham, Thomas Henighan, Adam Jermyn, Andy Jones, and 8
  others. 2025.
\newblock \href
  {https://transformer-circuits.pub/2025/attribution-graphs/biology.html} {On
  the biology of a large language model}.
\newblock \emph{Transformer Circuits Thread}.

\bibitem[{Luo et~al.(2025)Luo, Tan, Wong, Shi, Tang, Roongta, Cai, Luo, Li,
  Popa, and Stoica}]{deepscaler2025}
Michael Luo, Sijun Tan, Justin Wong, Xiaoxiang Shi, William~Y. Tang, Manan
  Roongta, Colin Cai, Jeffrey Luo, Li~Erran Li, Raluca~Ada Popa, and Ion
  Stoica. 2025.
\newblock Deepscaler: Surpassing o1-preview with a 1.5b model by scaling rl.
\newblock
  \url{https://pretty-radio-b75.notion.site/DeepScaleR-Surpassing-O1-Preview-with-a-1-5B-Model-by-Scaling-RL-19681902c1468005bed8ca303013a4e2}.
\newblock Notion Blog.

\bibitem[{Ma et~al.(2025{\natexlab{a}})Ma, He, Snell, Griggs, Min, and
  Zaharia}]{ma2025reasoning}
Wenjie Ma, Jingxuan He, Charlie Snell, Tyler Griggs, Sewon Min, and Matei
  Zaharia. 2025{\natexlab{a}}.
\newblock Reasoning models can be effective without thinking.
\newblock \emph{arXiv preprint arXiv:2504.09858}.

\bibitem[{Ma et~al.(2025{\natexlab{b}})Ma, He, Snell, Griggs, Min, and
  Zaharia}]{nothinking}
Wenjie Ma, Jingxuan He, Charlie Snell, Tyler Griggs, Sewon Min, and Matei
  Zaharia. 2025{\natexlab{b}}.
\newblock \href {https://arxiv.org/abs/2504.09858} {Reasoning models can be
  effective without thinking}.
\newblock \emph{Preprint}, arXiv:2504.09858.

\bibitem[{Muennighoff et~al.(2025)Muennighoff, Yang, Shi, Li, Fei-Fei,
  Hajishirzi, Zettlemoyer, Liang, Candès, and Hashimoto}]{S1RL}
Niklas Muennighoff, Zitong Yang, Weijia Shi, Xiang~Lisa Li, Li~Fei-Fei,
  Hannaneh Hajishirzi, Luke Zettlemoyer, Percy Liang, Emmanuel Candès, and
  Tatsunori Hashimoto. 2025.
\newblock \href {https://arxiv.org/abs/2501.19393} {s1: Simple test-time
  scaling}.
\newblock \emph{Preprint}, arXiv:2501.19393.

\bibitem[{OpenAI(2024)}]{o1_syscard}
OpenAI. 2024.
\newblock \href {https://cdn.openai.com/o1-system-card-20241205.pdf} {Openai o1
  system card}.

\bibitem[{OpenAI(2025)}]{o3o4_syscard}
OpenAI. 2025.
\newblock \href
  {https://cdn.openai.com/pdf/2221c875-02dc-4789-800b-e7758f3722c1/o3-and-o4-mini-system-card.pdf}
  {Openai o3 and o4-mini system card}.

\bibitem[{Qwen-Team(2024)}]{qwq-32b-preview}
Qwen-Team. 2024.
\newblock \href {https://qwenlm.github.io/blog/qwq-32b-preview/} {Qwq: Reflect
  deeply on the boundaries of the unknown}.

\bibitem[{Ramesh et~al.(2024)Ramesh, Hu, Chaimalas, Mehta, Sessa, Ammar, and
  Bogunovic}]{grpo}
Shyam~Sundhar Ramesh, Yifan Hu, Iason Chaimalas, Viraj Mehta, Pier~Giuseppe
  Sessa, Haitham~Bou Ammar, and Ilija Bogunovic. 2024.
\newblock \href {https://arxiv.org/abs/2405.20304} {Group robust preference
  optimization in reward-free rlhf}.
\newblock \emph{Preprint}, arXiv:2405.20304.

\bibitem[{Schulman et~al.(2017)Schulman, Wolski, Dhariwal, Radford, and
  Klimov}]{ppo}
John Schulman, Filip Wolski, Prafulla Dhariwal, Alec Radford, and Oleg Klimov.
  2017.
\newblock \href
  {http://dblp.uni-trier.de/db/journals/corr/corr1707.html#SchulmanWDRK17}
  {Proximal policy optimization algorithms.}
\newblock \emph{CoRR}, abs/1707.06347.

\bibitem[{Tang et~al.(2025)Tang, Zheng, Synnaeve, and
  Munos}]{optimizingforpassatk}
Yunhao Tang, Kunhao Zheng, Gabriel Synnaeve, and Rémi Munos. 2025.
\newblock \href {https://arxiv.org/abs/2503.19595} {Optimizing language models
  for inference time objectives using reinforcement learning}.
\newblock \emph{Preprint}, arXiv:2503.19595.

\bibitem[{Tian et~al.(2024)Tian, Peng, Song, Jin, Yu, Mi, and
  Yu}]{tian2024selfimprovementllmsimaginationsearching}
Ye~Tian, Baolin Peng, Linfeng Song, Lifeng Jin, Dian Yu, Haitao Mi, and Dong
  Yu. 2024.
\newblock \href {https://arxiv.org/abs/2404.12253} {{Toward Self-Improvement of
  LLMs via Imagination, Searching, and Criticizing}}.
\newblock \emph{Preprint}, arXiv:2404.12253.

\bibitem[{Valmeekam et~al.(2024)Valmeekam, Stechly, and
  Kambhampati}]{lrm-modulo}
Karthik Valmeekam, Kaya Stechly, and Subbarao Kambhampati. 2024.
\newblock \href {https://arxiv.org/abs/2409.13373} {Llms still can't plan; can
  lrms? a preliminary evaluation of openai's o1 on planbench}.
\newblock \emph{Preprint}, arXiv:2409.13373.

\bibitem[{Wang et~al.(2024)Wang, Wang, Athiwaratkun, Zhang, and
  Zou}]{wang2024mixtureofagentsenhanceslargelanguage}
Junlin Wang, Jue Wang, Ben Athiwaratkun, Ce~Zhang, and James Zou. 2024.
\newblock \href {https://arxiv.org/abs/2406.04692} {{Mixture-of-Agents Enhances
  Large Language Model Capabilities}}.
\newblock \emph{Preprint}, arXiv:2406.04692.

\bibitem[{Wang et~al.(2023)Wang, Xu, Lan, Hu, Lan, Lee, and
  Lim}]{planandsolvepromptingimprovingzeroshotchainofthoughtreasoningbylargelanguagemodels}
Lei Wang, Wanyu Xu, Yihuai Lan, Zhiqiang Hu, Yunshi Lan, Roy Ka-Wei Lee, and
  Ee-Peng Lim. 2023.
\newblock \href {https://arxiv.org/pdf/2305.04091} {Plan-and-solve prompting:
  Improving zero-shot chain-of-thought reasoning by large language models}.
\newblock \emph{arXiv preprint arXiv:2305.04091}.

\bibitem[{Wang et~al.(2022)Wang, Wei, Schuurmans, Le, Chi, Narang, Chowdhery,
  and Zhou}]{selfconsistencyimproveschainofthoughtreasoninginlanguagemodels}
Xuezhi Wang, Jason Wei, Dale Schuurmans, Quoc Le, Ed~H. Chi, Sharan Narang,
  Aakanksha Chowdhery, and Denny Zhou. 2022.
\newblock \href {https://arxiv.org/pdf/2203.11171} {Self-consistency improves
  chain of thought reasoning in language models}.
\newblock \emph{arXiv preprint arXiv:2203.11171}.

\bibitem[{Wang et~al.(2025)Wang, Liu, Xu, Liang, Chen, He, Song, Yu, Li, Zhang,
  Wang, Tu, Mi, and Yu}]{underthinking}
Yue Wang, Qiuzhi Liu, Jiahao Xu, Tian Liang, Xingyu Chen, Zhiwei He, Linfeng
  Song, Dian Yu, Juntao Li, Zhuosheng Zhang, Rui Wang, Zhaopeng Tu, Haitao Mi,
  and Dong Yu. 2025.
\newblock \href {https://arxiv.org/abs/2501.18585} {Thoughts are all over the
  place: On the underthinking of o1-like llms}.
\newblock \emph{Preprint}, arXiv:2501.18585.

\bibitem[{Wei et~al.(2022)Wei, Wang, Schuurmans, Bosma, brian ichter, Xia, Chi,
  Le, and Zhou}]{wei2022chain}
Jason Wei, Xuezhi Wang, Dale Schuurmans, Maarten Bosma, brian ichter, Fei Xia,
  Ed~H. Chi, Quoc~V Le, and Denny Zhou. 2022.
\newblock \href {https://openreview.net/forum?id=_VjQlMeSB_J} {{Chain of
  Thought Prompting Elicits Reasoning in Large Language Models}}.
\newblock In \emph{Advances in Neural Information Processing Systems}.

\bibitem[{Yao et~al.(2023)Yao, Yu, Zhao, Shafran, Griffiths, Cao, and
  Narasimhan}]{treeofthoughtsdeliberateproblemsolvingwithlargelanguagemodels}
Shunyu Yao, Dian Yu, Jeffrey Zhao, Izhak Shafran, Tom Griffiths, Yuan Cao, and
  Karthik Narasimhan. 2023.
\newblock \href
  {https://proceedings.neurips.cc/paper_files/paper/2023/file/271db9922b8d1f4dd7aaef84ed5ac703-Paper-Conference.pdf}
  {{Tree of Thoughts: Deliberate Problem Solving with Large Language Models}}.
\newblock In \emph{Advances in Neural Information Processing Systems},
  volume~36, pages 11809--11822. Curran Associates, Inc.

\bibitem[{Yeo et~al.(2025)Yeo, Tong, Niu, Neubig, and
  Yue}]{demystifyinglongchainofthoughtreasoning}
Edward Yeo, Yuxuan Tong, Morry Niu, Graham Neubig, and Xiang Yue. 2025.
\newblock \href {https://arxiv.org/abs/2502.03373} {Demystifying long
  chain-of-thought reasoning in llms}.
\newblock \emph{Preprint}, arXiv:2502.03373.

\bibitem[{Yue et~al.(2025)Yue, Chen, Lu, Zhao, Wang, Song, and
  Huang}]{yue2025does}
Yang Yue, Zhiqi Chen, Rui Lu, Andrew Zhao, Zhaokai Wang, Shiji Song, and Gao
  Huang. 2025.
\newblock Does reinforcement learning really incentivize reasoning capacity in
  llms beyond the base model?
\newblock \emph{arXiv preprint arXiv:2504.13837}.

\bibitem[{Zheng et~al.(2025)Zheng, Xie, Wang, and
  Hooi}]{LLMmctsAutomaticHeuristicDesign}
Zhi Zheng, Zhuoliang Xie, Zhenkun Wang, and Bryan Hooi. 2025.
\newblock \href {https://arxiv.org/abs/2501.08603} {{Monte Carlo Tree Search
  for Comprehensive Exploration in LLM-Based Automatic Heuristic Design}}.
\newblock \emph{Preprint}, arXiv:2501.08603.

\end{thebibliography}

\end{document}